\documentclass[lettersize,journal]{IEEEtran}
\usepackage{amsmath,amsfonts}
\usepackage{algorithm}
\usepackage{array}
\usepackage[caption=false,font=normalsize,labelfont=sf,textfont=sf]{subfig}
\usepackage{textcomp}
\usepackage{stfloats}
\usepackage{url}
\usepackage{verbatim}
\usepackage{graphicx}
\usepackage{cite}
\usepackage{booktabs}
\usepackage{amssymb}
\usepackage{multirow}
\usepackage{algorithm}
\usepackage{algpseudocode}

\algrenewcommand\algorithmicrequire{\textbf{Input:}}
\algrenewcommand\algorithmicensure{\textbf{Output:}}

\usepackage[colorlinks=true,
            linkcolor=blue,
            citecolor=black,
            urlcolor=blue]{hyperref}

\begin{document}

\title{NeSy-CSA: A Neuro-Symbolic Framework for Open-Ended Critical Scenario Attribution
}

\author{Qitong Chu$^1$, Xunjie He$^1$, Chen Deng$^1$, Huaxin Pei$^{2}$, Yufeng Yue$^{1*}$

\thanks{This work was supported in part by the National Natural Science Foundation
of China under Grants 62473050. \textit{(Corresponding Author: Yufeng Yue.)}}
 

\thanks{$^1$Qitong Chu, Xunjie He, Chen Deng and Yufeng Yue are with the School of Automation, Beijing Institute of Technology, Beijing 100081, China.(email: chuqitong@bit.edu.cn, 3120215467@bit.edu.cn, 1120232687@bit.edu.cn, yueyufeng@bit.edu.cn)}
\thanks{$^2$ Huaxin Pei is with the Department of Automation, Tsinghua University, Beijing 100084, China. (email: 
phx17@tsinghua.org.cn). }

}



\maketitle

\begin{abstract}

Understanding why discovered scenarios become critical in scenario-based testing is essential for effectively leveraging them in decision-making systems. Reasoning about such criticality can be formulated as an attribution problem. However, across different decision-making tasks, the causes of criticality may involve diverse state variables, interaction patterns, and failure mechanisms, making attribution an inherently open-ended problem beyond predefined explanation spaces. Existing attribution methods still struggle to balance open-ended reasoning flexibility with the interpretability and traceability required for critical scenario reasoning. To address this limitation, we propose NeSy-CSA, a neuro-symbolic framework that transforms open-ended critical scenario attribution from unconstrained explanation generation into structured and traceable reasoning. NeSy-CSA narrows the attribution space by selecting relevant factors, makes the reasoning process traceable through a dependency-aware evidence graph, and executes symbolic reasoning procedures derived from atomic operations, coordinated with evidence-constrained neural inference to support flexible open-ended attribution. We further introduce a process-level and result-level assessment module to evaluate the structural validity of the attribution process and the behavioral effectiveness of the attribution results under controlled interventions. Experiments across four decision-making environments show that NeSy-CSA improves two intervention-based measures of attribution effectiveness by 18.32\% and 13.67\% over LLM-based baselines. These results demonstrate its potential to transform discovered critical scenarios into reusable knowledge for subsequent testing and safety analysis.



\end{abstract}


\section{Introduction}

\IEEEPARstart{D}{ecision-making} agents, such as those in robots \cite{zhu2025confidence,chi2026dynamic} and autonomous vehicles \cite{taghavifar2025behaviorally}, are increasingly deployed in real-world systems, where safety is essential for trustworthy operation.  Scenario-based testing has therefore become an important paradigm for assessing these agents before deployment \cite{jiang2025hybrid,ji2025autonomous,weng2023comparability}, because it can proactively generate rare yet critical scenarios that may trigger failures or unsafe outcomes. However, discovering critical scenarios alone is not enough. This highlights the need to understand why a scenario is critical, thereby supporting more informative testing and more effective safety analysis.


Understanding why a scenario is critical is, in essence, an attribution problem. Traditional attribution methods, such as fault diagnosis \cite{gao2015survey} and causal reasoning \cite{pearl2009causality}, rely on predefined variables or reasoning rules. This makes their inference stable and interpretable, but also ties the attribution process to a fixed analysis space. Recently, large language models (LLMs) offer a more flexible attribution paradigm \cite{ achiam2023gpt,liu2024deepseek}, where attribution reasoning can be adapted to task-specific instructions and different scenario evidence without relying on a fully predefined explanation space. However, their reasoning process is largely implicit and difficult to trace, making the resulting attribution conclusions hard to verify and prone to hallucination. 
\IEEEpubidadjcol

\begin{figure}[t] 
    \centering
    \includegraphics[width=0.95\linewidth]{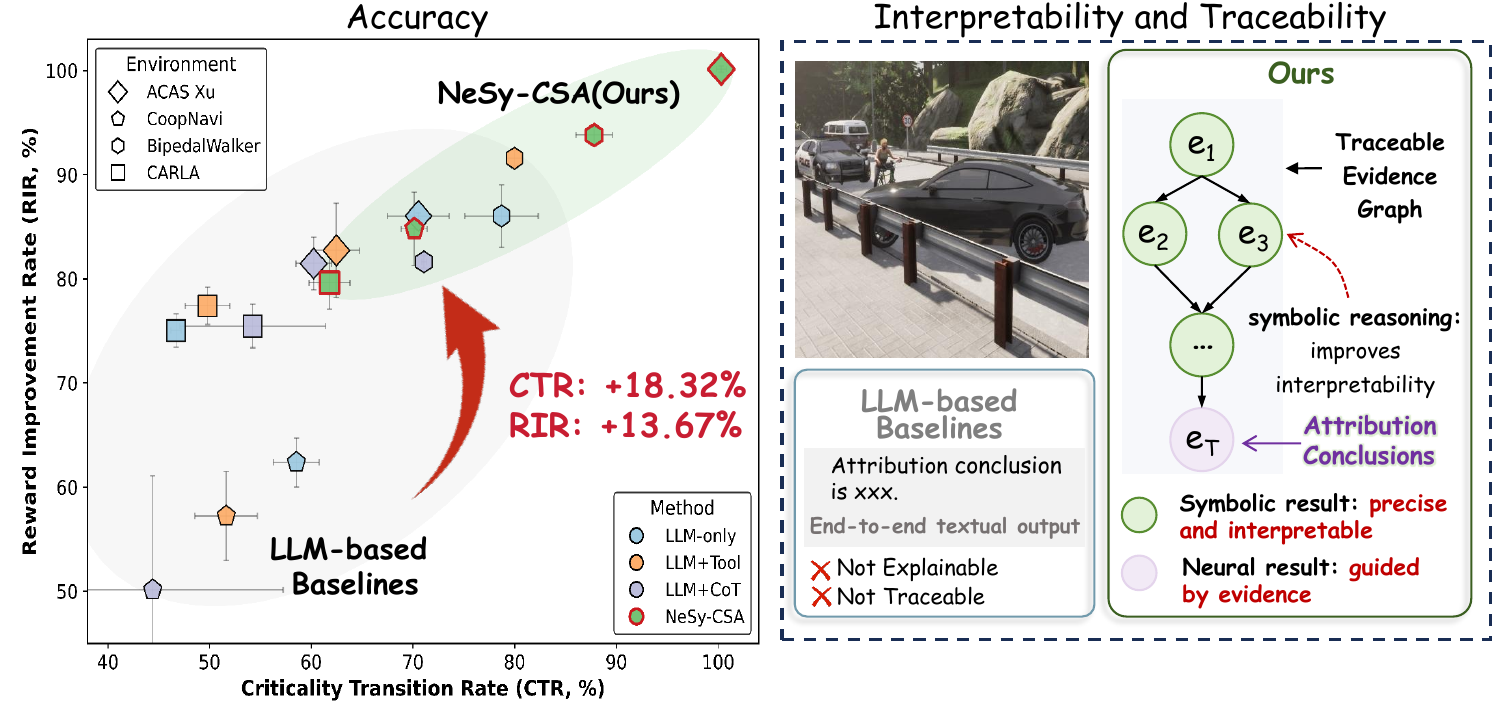} %
    \caption{NeSy-CSA improves critical scenario attribution over LLM-based baselines while providing interpretable and traceable reasoning. The left panel compares attribution effectiveness across four environments, and the right panel illustrates how symbolic and neural evidence is organized into a traceable attribution process.}
    \label{fig:nesy_csa_intro}
\vspace{-0.5cm}
\end{figure}


Neuro-symbolic reasoning has emerged as a promising paradigm to bridge this gap. Its core idea is to combine neural models for interpreting unstructured semantics, while symbolic components enforce explicit reasoning constraints to ensure interpretable inference \cite{li2025survey}. This integration enables both flexible semantic reasoning and interpretable inference. Such capabilities have been demonstrated in domains such as robot task planning \cite{du2026fast,silver2023predicate,hansen2022bisimulation,chitnis2022learning} and safe reinforcement learning \cite{sharifi2023towards,acharya2023neurosymbolic}, where symbolic constraints are used to guide neural policy learning and verify learned policies against formal specifications. However, the effectiveness of existing neuro-symbolic methods often relies on predefined symbolic knowledge, which constrains inference to explicitly encoded knowledge and limits adaptability to new tasks. This limitation makes them mainly suitable for settings with clearly specified objectives, rendering them difficult to apply directly to critical scenario attribution, where no gold-standard answer exists. Consequently, a fundamental challenge remains: can we achieve cross-task critical scenario attribution in a way that remains interpretable and reliably grounded in evidence?


Answering this question requires a way to constrain open-ended attribution instead of reducing it to a fixed reasoning template. However, this faces the following difficulties: \textbf{i)} Relevant attribution factors must be selected from an open-ended candidate set, where LLM-generated factors may repeat similar meanings or show weak relevance to the observed critical behavior. A statistically grounded filtering mechanism is therefore needed to retain a compact set of representative attribution dimensions. \textbf{ii)} The reasoning process must be traceable, whereas end-to-end attribution generation does not expose how the conclusion is derived. 
The attribution objective should thus be decomposed into predecessor-dependent subtasks, making the reasoning procedure explicit and structured. \textbf{iii)} The reasoning mechanism must be interpretable yet not limited to a fixed symbolic rule base, and flexible yet not reduced to unconstrained LLM inference. In open-ended attribution, formalizable reasoning steps vary across tasks and failure mechanisms, while semantic judgments must be grounded in intermediate evidence. This motivates a neuro-symbolic execution mechanism that dynamically constructs explicit symbolic procedures for formalizable subtasks, without relying on a comprehensive predefined rule base, while applying evidence-constrained LLM inference to semantic or uncertain subtasks. Together, these requirements constrain attribution at the level of relevant factors and reasoning process, while grounding all steps in intermediate evidence, transforming unconstrained explanation generation into interpretable and traceable reasoning.

Based on this insight, we propose \textbf{NeSy-CSA} (\textbf{Neuro-Symbolic Critical Scenario Attribution}), a neuro-symbolic framework for open-ended critical scenario attribution across decision-making agents. Fig.~\ref{fig:nesy_csa_intro} contrasts NeSy-CSA with LLM-based attribution in terms of attribution effectiveness, interpretability, and traceability. NeSy-CSA consists of three main modules: refining candidate factors into a compact set of key attribution factors, constructing a predecessor-dependent subtask graph to organize intermediate evidence, and executing the graph through a hybrid neuro-symbolic mechanism. In this mechanism, symbolic reasoning is dynamically generated on demand from subtask semantics using lightweight atomic functions, without relying on a large predefined symbolic rule base, while semantic or uncertain subtasks are handled via evidence-constrained LLM inference. Through this design, NeSy-CSA transforms critical scenario attribution from implicit end-to-end generation into an interpretable and traceable reasoning process while preserving cross-task flexibility. Since open-ended attribution lacks standard answers, its evaluation is inherently challenging and cannot rely on conventional reference-based metrics. We therefore introduce a two-level evaluation that considers both process validity and result effectiveness.

In summary, we formulate critical scenario attribution as an open-ended reasoning problem for decision-making agents and propose NeSy-CSA. Our contributions are as follows:

i) We develop a key factor construction method that builds a compact attribution space through knowledge-guided generation, semantic deduplication, and data-driven filtering.

ii) We introduce a predecessor-dependent subtask graph to organize intermediate evidence for traceable attribution.

iii) We propose a neuro-symbolic hybrid reasoning strategy, where formalizable subtasks are handled by dynamically generated symbolic operations, while semantic subtasks are addressed by neural reasoning guided by intermediate evidence, enabling flexible and interpretable reasoning.

iv) We design a two-level evaluation method to assess both the structural validity of the attribution process and the effectiveness of the attribution results.

The remainder of this paper is structured as follows. Section II summarizes the related studies on scenario-based testing and attribution reasoning. Section III introduces the proposed NeSy-CSA framework, including key attribution factor refinement, predecessor-dependent subtask graph construction, neuro-symbolic hybrid reasoning, and the evaluation strategy for open-ended attribution. Section IV presents the experimental settings and discusses the results. Section V concludes this paper and discusses future research directions.

\section{Related Work}

\subsection{Scenario-Based Testing and Critical Scenario Generation}

Scenario-based testing evaluates decision-making agents under varied environmental and interaction conditions, aiming to reveal rare but critical behaviors that may only emerge in specific scenarios. Existing studies have addressed this problem through several technical lines. Fuzzing-based methods adapt the fuzzing paradigm to decision-making agents by iteratively selecting scenario seeds, mutating them, executing the agent, and prioritizing subsequent tests according to task-guided signals ~\cite{pang2022mdpfuzz,wang2023fuzzing,chu2025dicritest}. While these signals guide local exploration and increase the chance of reaching critical behaviors, the resulting perturbations are largely random and do not capture reusable patterns underlying critical outcomes. Objective-guided scenario search methods formulate the search for critical scenarios as a goal-directed testing problem. They define explicit risk or adversarial objectives and optimize scenario parameters or agent behaviors to increase the probability of critical outcomes~\cite{zolfagharian2023search,feng2023dense,weng2023comparability}. 
Generative-model-based methods learn scenario distributions from data or simulation interactions, enabling new scenarios beyond manually designed templates~\cite{li2023generative}. Recent diffusion-based approaches further steer this generation toward realistic and safety-critical cases~\cite{xu2025diffscene}. LLM-driven scenario generation methods exploit the semantic understanding and world knowledge of large language models to translate high-level testing requirements into complex scenario descriptions or executable testing scenarios~\cite{li2024chatgpt,lu2025omnitester}. 

While these methods improve exploration of large and complex scenario spaces, they focus mainly on discovering critical scenarios and provide limited understanding of why a scenario is critical. Consequently, the discovered cases often remain isolated and provide little guidance for subsequent scenario generation or safety improvement. In contrast, this paper focuses on critical scenario attribution, aiming to identify interpretable factors behind such scenarios and make the reasoning process traceable, so that the discovered cases can be transformed into reusable knowledge for subsequent testing.


\subsection{Attribution and Reasoning Methods}


Critical scenario attribution aims to explain why a scenario leads to task failure, which requires reasoning over scenario evidence to connect observed behaviors with possible causes.  This subsection reviews three related lines of methods: traditional analytical methods, LLM-based methods, and neuro-symbolic methods.

Traditional attribution methods offer structured ways to relate abnormal outcomes to potential causes, using system models, signal patterns, or diagnostic assumptions~\cite{reiter1987theory,gao2015survey}. Causal and counterfactual reasoning examines whether changes to specific variables would alter observed outcomes~\cite{pearl2009causality,halpern2005causes,madumal2020explainable}. While these methods provide interpretable analyses, they generally assume that relevant variables, system structures, or intervention spaces are predefined. As a result, they cannot handle open-ended scenario evidence expressed in natural language or other unstructured forms.

LLM-based methods provide a flexible way to handle open-ended attribution, because large language models can draw on broad pretrained knowledge and adapt to diverse reasoning tasks from natural-language instructions~\cite{achiam2023gpt,guo2025deepseek}. Given heterogeneous inputs and task specifications, LLMs can interpret multi-source evidence and reason about potential causal relations across different environments. 
Chain-of-thought prompting enables structured reasoning in complex tasks by explicitly decomposing the inference process into a sequence of intermediate steps~\cite{wei2022chain}, thereby improving the transparency of intermediate reasoning.
Tool-augmented reasoning further extends LLM capabilities by allowing the invocation of external tools for computation and information retrieval~\cite{yao2023react,schick2023toolformer}. A representative direction is retrieval-augmented generation (RAG), which enhances reasoning by incorporating external knowledge sources\cite{lewis2020retrieval,liu2023evaluating}. While such retrieval-based methods can improve factual grounding and reduce hallucination, their reliability is still dependent on the coverage and authority of the underlying data sources, especially when web-based content is involved.

To further improve the reliability and interpretability of reasoning, neuro-symbolic methods combine the flexibility of neural models with explicit symbolic procedures~\cite{pan2023logic,bhuyan2024neuro}. Recent studies have applied this paradigm to task planning and safe reinforcement learning, where learned models are guided by symbolic constraints to improve reliability and interpretability~\cite{chi2026instructflow,wang2024imperative}. Related efforts have also explored how symbolic procedures can support explicit reasoning in neural systems. For example, Prenosil et al. proposed a neuro-symbolic method for rule-guided clinical label inference, where an LLM identifies report-level evidence from free-text medical reports and a symbolic expert system reasons over this evidence to derive labels~\cite{prenosil2025neuro}. Similarly, NeSyPr uses neuro-symbolic methods for embodied reasoning, where symbolic knowledge provides structured task steps that guide LLM-based agents in inferring actions for multi-step tasks~\cite{choi2026nesypr}. 
However, their reasoning ability still depends on symbolic knowledge specified before deployment. This limits their direct use in critical scenario attribution, where the causes to be explained are unknown in advance and the attribution space remains open.

\begin{figure*}[t]
  \centering
  \includegraphics[width=\textwidth]{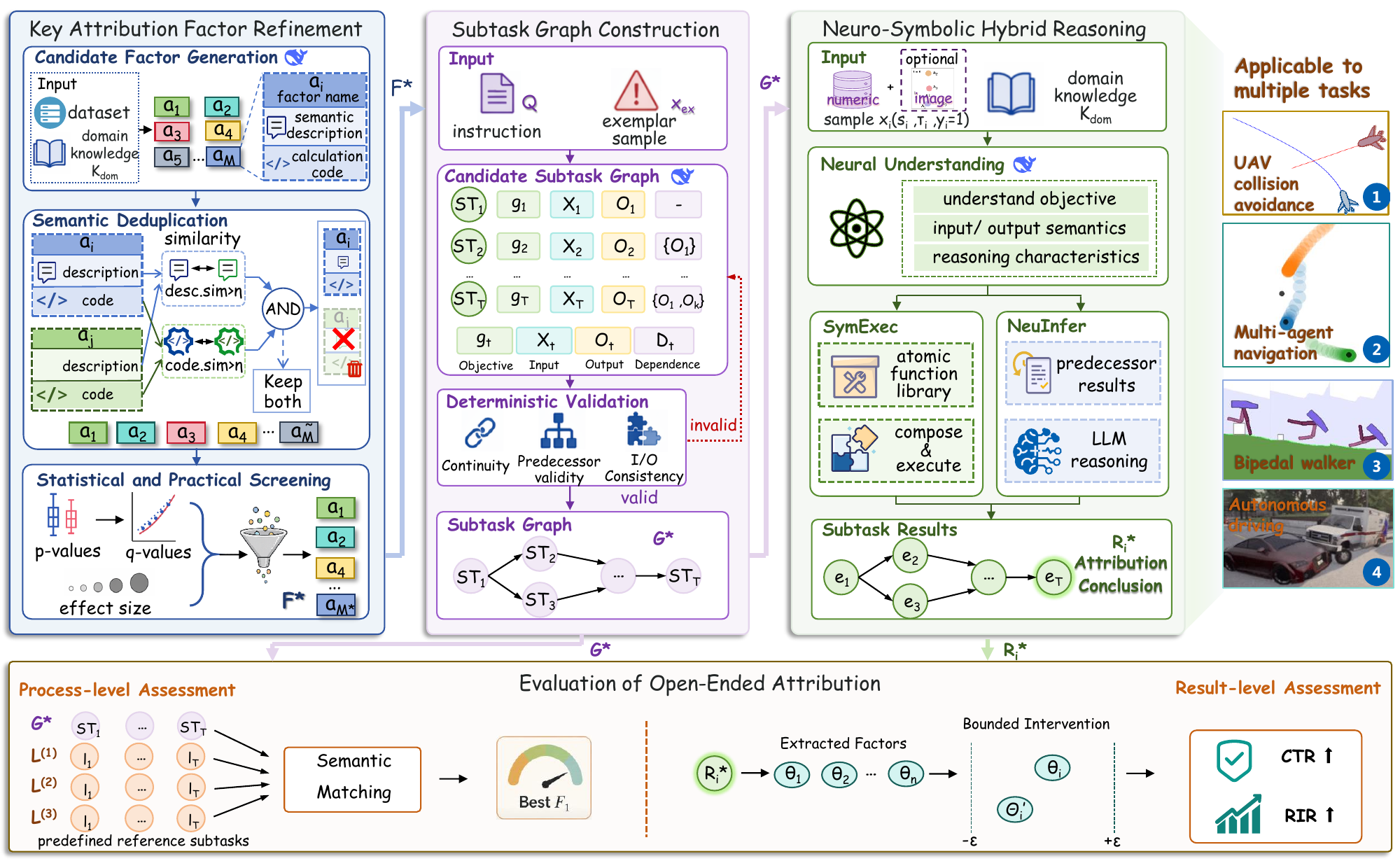}
  \caption{\textbf{Overview of the NeSy-CSA framework.} The pipeline includes key attribution factor refinement, subtask graph construction, and neuro-symbolic hybrid reasoning that combines symbolic computation for formalizable subtasks with neural inference for the remaining subtasks. The resulting attribution is evaluated from both process-level and result-level perspectives.}
  \label{fig:fig2}
  \vspace{-0.5cm}
\end{figure*}

\section{Methodology}

In this section, we present NeSy-CSA, a neuro-symbolic framework for open-ended critical scenario attribution in decision-making agents. The framework first refines open-ended candidate factors into a compact set of informative attribution dimensions, then decomposes the attribution objective into a predecessor-dependent subtask graph. Based on this graph, formalizable subtasks are handled by symbolic procedures, while semantic or uncertain subtasks are handled by neural reasoning. Finally, the attribution results are evaluated from both process-level and result-level perspectives.

\subsection{Problem Formulation and Method Overview}

The objective of critical scenario attribution is to determine why a discovered scenario becomes critical and to produce attribution results. NeSy-CSA formulates this problem as a structured reasoning process: it refines candidate attribution factors, decomposes the attribution objective into dependent subtasks, and assigns each subtask to neural or symbolic reasoning according to its reasoning characteristics.

Formally, let a critical sample be denoted by $x_i=(s_i,\tau_i,y_i)$, where $s_i$ denotes the initial scenario information, including structured state parameters and, when available, an initial scenario image, $\tau_i$ is the interaction trajectory, and $y_i=1$ indicates that the execution leads to a critical outcome, such as failure or unsafe behavior. The historical dataset is denoted by
\begin{equation}
\mathcal{D}=\{(s_j,\tau_j,y_j)\}_{j=1}^{N},
\end{equation}
where $y_j\in\{0,1\}$ distinguishes critical samples from non-critical ones. Given the target critical sample $x_i$, the historical dataset $\mathcal{D}$, the attribution instruction $Q$, and domain prior knowledge $K_{\mathrm{dom}}$, NeSy-CSA produces
\begin{equation}
z_i=\langle F^{*}, G^{*}, R_i^{\star}\rangle
= \mathcal{M}(x_i,\mathcal{D},Q;K_{\mathrm{dom}}),
\end{equation}
where $F^{*}$ is the refined set of key attribution factors, $G^{*}$ is the predecessor-dependent subtask graph, and $R_i^{\star}$ is the final attribution conclusion.


A key challenge in critical scenario attribution is that the causes of criticality are task dependent and cannot be fully specified in advance. Existing neuro-symbolic reasoning with a fixed symbolic knowledge base is therefore insufficient, as it cannot predefine all required rules or fully formalize attribution problems without standard answers. Meanwhile, unconstrained open-ended attribution may introduce redundant factors, skip intermediate evidence, or produce conclusions that are difficult to inspect. NeSy-CSA addresses these limitations through factor refinement, constructing a predecessor-dependent subtask graph, and neuro-symbolic attribution, which jointly narrow the attribution space, organize the reasoning workflow, and combine symbolic reasoning with neural reasoning guided by intermediate evidence:
\begin{equation}
\mathcal{X}
\xrightarrow{\ \Psi_{F}\ }
\mathcal{X}_{F}
\xrightarrow{\ \Psi_{G}\ }
\mathcal{X}_{G}
\xrightarrow{\ \Psi_{R}\ }
\mathcal{X}_{R}
\xrightarrow{\ \omega\ }
\{R_i^{\star}\}.
\end{equation}
Here, $\Psi_{F}$ denotes the factor-level constraint that refines open-ended candidate factors into key attribution factors, $\Psi_{G}$ denotes the structure-level constraint that organizes attribution into a predecessor-dependent subtask graph, $\Psi_{R}$ denotes the reasoning-level constraint imposed during neuro-symbolic hybrid execution, and $\omega$ maps the terminal reasoning output to the final attribution conclusion.


Since critical scenario attribution is open-ended and often lacks standard answers, reference-based evaluation alone is insufficient. NeSy-CSA therefore evaluates attribution along two complementary dimensions: the structural validity of the reasoning workflow and the practical effectiveness of the identified factors. The overall framework, including the attribution modules and evaluation strategy, is illustrated in Fig.~\ref{fig:fig2}, and the attribution process is summarized in Algorithm~\ref{alg:nesy_csa}.

\begin{algorithm}[t]
\caption{NeSy-CSA for Critical Scenario Attribution}
\label{alg:nesy_csa}
\begin{algorithmic}[1]
\Require Critical sample $x_i=(s_i,\tau_i,y_i=1)$, dataset $\mathcal{D}$, instruction $Q$, prior knowledge $K_{\mathrm{dom}}$, atomic library $\mathcal{A}$
\Ensure Attribution result $R_i^{\star}$

\State $F^{\mathrm{cand}} \gets \textsc{GenerateFactors}(\mathcal{D},K_{\mathrm{dom}})$
\State $\widetilde{F}^{\mathrm{cand}} \gets \textsc{SemanticDedup}(F^{\mathrm{cand}})$
\State $F^{*} \gets \textsc{DataDrivenFilter}(\widetilde{F}^{\mathrm{cand}}, \mathcal{D}, \alpha, \delta)$
\Comment{Factor refinement}

\State $x_{\mathrm{ex}} \gets \textsc{SampleCritical}(\mathcal{D})$
\Repeat
    \State $\hat{G} \gets \textsc{Decompose}(F^{*},Q,x_{\mathrm{ex}})$
    \State valid $\gets \textsc{ValidateGraph}(\hat{G})$
\Until{valid}
\State $G^{*} \gets \hat{G}$
\Comment{Subtask graph construction}

\For{each $ST_t \in G^{*}$ in dependency order}
    \State $E_t \gets \{e_j \mid j \in D_t\}$
    \If{\textsc{Formalizable}$(ST_t)$ and \textsc{Verifiable}$(ST_t)$}
        \State $\Pi_t \gets \textsc{Instantiate}(ST_t,\mathcal{A})$
        \State $e_t \gets \Pi_t(s_i,\tau_i)$
    \Else
        \State $e_t \gets \textsc{NeuInfer}(ST_t,s_i,\tau_i,E_t,K_{\mathrm{dom}})$
    \EndIf
\EndFor
\State $R_i^{\star} \gets e_T$
\State \Return $R_i^{\star}$
\Comment{Neuro-symbolic hybrid reasoning}
\end{algorithmic}
\end{algorithm}

\subsection{Key Attribution Factor Refinement}

The first stage of NeSy-CSA aims to construct a compact and reliable attribution space from the historical dataset $\mathcal{D}$. This module is needed because open-ended attribution cannot rely on a fixed factor set, while directly using all LLM-generated candidates may introduce redundancy and noise into subsequent reasoning. Although knowledge-guided generation can cover diverse potential causes, the generated candidates may contain overlapping semantics, similar extraction procedures, weak relevance, or spurious correlations, which can unnecessarily enlarge the attribution space and distract subsequent subtask graph construction and neuro-symbolic attribution.

To address this issue, NeSy-CSA constructs key attribution factors through three steps: knowledge-guided generation, semantic deduplication, and data-driven filtering. Knowledge-guided generation expands the candidate space beyond manually predefined factors; semantic deduplication removes conceptually or operationally similar candidates; and data-driven filtering retains factors with reliable and practically meaningful differences between critical and non-critical samples. Together, these steps preserve the coverage of open-ended factor generation while producing a focused and empirically supported factor set for subsequent attribution reasoning.

\paragraph{Knowledge-guided Generation}
Conditioned on the historical dataset $\mathcal{D}$ and a lightweight domain prior $K_{\mathrm{dom}}$, the LLM generates a candidate factor set
\begin{equation}
F^{\mathrm{cand}}=\{a_{1},a_{2},\dots,a_{M}\},
\end{equation}
corresponding to the \textsc{GenerateFactors} step in Algorithm~\ref{alg:nesy_csa}. Each candidate factor $a_m$ is associated with an executable extractor $\psi_m$, which converts the factor into a computable variable over historical interaction data. For a sample $x_j=(s_j,\tau_j,y_j)$, the value of the $m$-th candidate factor is defined as

\begin{equation}
u_{j,m}=\psi_m(s_j,\tau_j).
\end{equation}

\paragraph{Semantic Deduplication}

LLM-based factor generation may introduce redundant candidates due to the surface variability of natural language generation: factors with different names or descriptions may still refer to similar attribution concepts or rely on nearly identical extraction procedures. Therefore, NeSy-CSA performs semantic deduplication before statistical screening to avoid treating such duplicate candidates as independent evidence. For each pair of candidate factors, similarity is measured from two complementary perspectives: i) the factor-level description, constructed from the factor name and its textual explanation, and ii) the corresponding extraction procedure. The description-level similarity captures conceptual overlap between two factors, whereas the procedure-level similarity captures whether they are operationalized in a similar way. A pair of factors is regarded as redundant only when both similarities exceed the deduplication threshold $\eta_d$. In this case, the redundant counterpart is removed, preventing repeated or nearly equivalent factors from being tested multiple times in the subsequent statistical analysis. Formally, this operation corresponds to \textsc{SemanticDedup} in Algorithm~\ref{alg:nesy_csa}, which maps the original candidate set to the deduplicated candidate set:

\begin{equation}
\widetilde{F}^{\mathrm{cand}}=\{\tilde{a}_{1},\tilde{a}_{2},\dots,\tilde{a}_{\widetilde{M}}\},
\end{equation}
where $\widetilde{M}\le M$. Each retained factor $\tilde{a}_m\in\widetilde{F}^{\mathrm{cand}}$ inherits the executable extractor of the surviving candidate after deduplication. Unless otherwise specified, the index $m$ in the following analysis refers to factors in $\widetilde{F}^{\mathrm{cand}}$. 

\paragraph{Data-driven Filtering}
Even after deduplication, factors may not distinguish critical from non-critical samples. Therefore, statistical screening is further performed on $\widetilde{F}^{\mathrm{cand}}$ to identify factors whose values differ between the two groups. Specifically, the historical dataset $\mathcal{D}$ is divided into $\mathcal{D}^{nc}=\{x_j\in\mathcal{D}\mid y_j=0\}$ and $\mathcal{D}^{c}=\{x_j\in\mathcal{D}\mid y_j=1\}$, denoting the non-critical and critical sets, respectively. Welch's $t$-test is adopted to assess whether the mean factor values differ between the two groups. This choice is appropriate because the two groups may be imbalanced and cannot be assumed to have equal variances. In addition, although the raw factor values may not strictly follow a normal distribution, the relatively large sample size allows the sampling distribution of the mean to be approximated as normal according to the Central Limit Theorem, making the test reasonably robust to moderate deviations from normality. For the $m$-th factor in the deduplicated candidate set $\widetilde{F}^{\mathrm{cand}}$, the Welch statistic is computed by comparing its values on $\mathcal{D}^{c}$ and $\mathcal{D}^{nc}$ as follows:

\begin{equation}
t_m=
\frac{\bar{u}_{m}^{\mathrm{c}}-\bar{u}_{m}^{\mathrm{nc}}}
{\sqrt{\frac{(s_{m}^{\mathrm{nc}})^2}{n^{\mathrm{nc}}}+\frac{(s_{m}^{\mathrm{c}})^2}{n^{\mathrm{c}}}}},
\end{equation}
where $\bar{u}_{m}^{\mathrm{nc}}$, $s_{m}^{\mathrm{nc}}$, and $n^{\mathrm{nc}}$ denote the sample mean, sample standard deviation, and sample size of the $m$-th factor on $\mathcal{D}^{\mathrm{nc}}$, respectively, and $\bar{u}_{m}^{\mathrm{c}}$, $s_{m}^{\mathrm{c}}$, and $n^{\mathrm{c}}$ are defined analogously on $\mathcal{D}^{\mathrm{c}}$. The corresponding two-sided $p$-value, denoted by $p_m$, is then computed under the Student's $t$ distribution with Welch's degrees of freedom:
\begin{equation}
\nu_m=
\frac{\left(\frac{(s_m^{nc})^2}{n^{nc}}+\frac{(s_m^{c})^2}{n^{c}}\right)^2}
{\frac{\left((s_m^{nc})^2/n^{nc}\right)^2}{n^{nc}-1}
+
\frac{\left((s_m^{c})^2/n^{c}\right)^2}{n^{c}-1}}.
\end{equation}
Since multiple candidate factors are tested simultaneously, directly thresholding the raw $p$-values may increase the probability of false discoveries. To mitigate this issue, the Benjamini--Hochberg (BH) procedure is applied over the deduplicated candidate set. Let
\begin{equation}
p_{(1)} \leq p_{(2)} \leq \cdots \leq p_{(\widetilde{M})}
\end{equation}
denote the raw $p$-values sorted in ascending order, where $p_{(1)}$ is the smallest and $p_{(\widetilde{M})}$ is the largest. The largest index $k$ satisfying
\begin{equation}
p_{(k)} \leq \frac{k}{\widetilde{M}}\alpha
\end{equation}
is identified, where $\alpha$ is the target false discovery rate. Candidate factors whose raw $p$-values satisfy $p_m \leq p_{(k)}$ are retained as statistically significant after correction. Let $\tilde{p}_m$ denote the BH-adjusted $p$-value of the $m$-th candidate factor. In this way, the screening procedure controls false discoveries caused by repeated hypothesis testing while retaining factors with statistically reliable group differences.

Although BH correction identifies factors with statistically reliable group differences, statistical significance does not necessarily imply a meaningful difference in magnitude. In large datasets, a small numerical difference between critical and non-critical samples may still yield a significant $p$-value, making it insufficient to rely on significance testing alone. To further assess the practical magnitude of the difference, Hedges' $g$ is computed for each retained factor:

\begin{equation}
g_m^{\mathrm{hedge}}
=
J(N_m)\cdot
\frac{\bar{u}_{m}^{c}-\bar{u}_{m}^{nc}}
{s_{m}^{\mathrm{pool}}},
\end{equation}
where
\begin{equation}
s_{m}^{\mathrm{pool}}
=
\sqrt{
\frac{(n^{c}-1)(s_{m}^{c})^2+(n^{nc}-1)(s_{m}^{nc})^2}
{n^{c}+n^{nc}-2}
},
\end{equation}
\begin{equation}
N_m=n^{c}+n^{nc},
\qquad
J(N_m)=1-\frac{3}{4N_m-9}.
\end{equation}
Here, $s_{m}^{\mathrm{pool}}$ is the pooled standard deviation, $N_m$ is the total number of samples involved in the comparison, and $J(N_m)$ is the small-sample correction term. The correction term makes Hedges' $g$ a bias-adjusted standardized measure of group difference, which is preferable when the two groups have unequal sample sizes.

The above statistical significance test, BH correction, and effect-size filtering together form the data-driven filtering operation, denoted as \textsc{DataDrivenFilter} in Algorithm~\ref{alg:nesy_csa}. Applying this operation to the deduplicated candidate set gives the final key factor set:
\begin{equation}
F^{*}
=
\left\{
\tilde{a}_m\in \widetilde{F}^{\mathrm{cand}}
\;\middle|\;
\tilde{p}_{m}<\alpha,\;
\left|g_{m}^{\mathrm{hedge}}\right|>\delta
\right\},
\end{equation}
where $\alpha$ is the significance threshold and $\delta$ is the minimum effect-size threshold. This criterion ensures that the retained factors are not only statistically distinguishable between non-critical and critical samples, but also practically informative for subsequent attribution reasoning. 

After refinement, the original open-ended explanation space is restricted to the factor-constrained subspace
\begin{equation}
\mathcal{X}_{F}
=
\{h\in\mathcal{X}\mid \mathrm{Var}(h)\subseteq F^{*}\},
\end{equation}
where $\mathcal{X}$ denotes the unconstrained explanation space, $\mathcal{X}_{F}$ is the factor-refined explanation space, $h$ is a candidate attribution hypothesis, and $\mathrm{Var}(h)$ denotes the set of attribution factors referenced by $h$. In this way, subsequent reasoning is guided toward a smaller and more reliable factor space, thereby reducing semantic noise while preserving the most informative attribution dimensions.

\subsection{Predecessor-Dependent Subtask Graph Construction}

\begin{figure}[t]
  \centering
  \includegraphics[width=0.95\columnwidth]{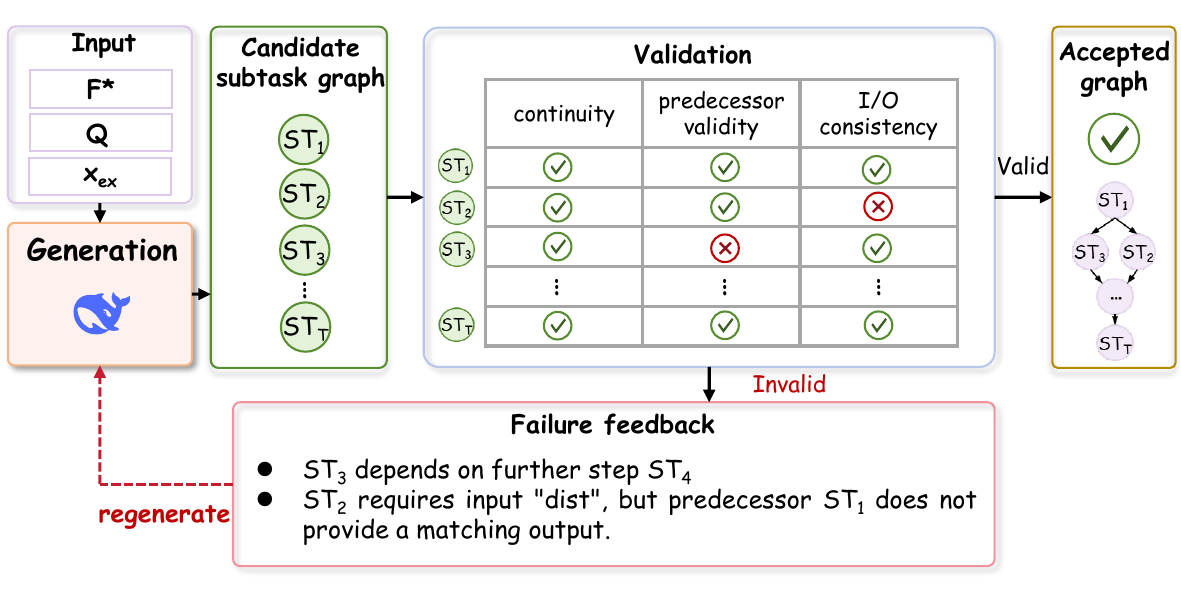}
\caption{Generation and validation of NeSy-CSA's predecessor-dependent subtask graph.}
  \label{fig:subtask_graph}
  \vspace{-0.5cm}
\end{figure}

The second stage of NeSy-CSA organizes open-ended attribution into a traceable predecessor-dependent subtask graph. Directly generating an attribution conclusion in an end-to-end manner may skip intermediate evidence and make the reasoning process difficult to inspect. In addition, generating a separate reasoning graph for every critical sample would increase inference cost and introduce unnecessary variation in subtask organization. NeSy-CSA therefore constructs a task-specific predecessor-dependent subtask graph once for each task. The graph is generated from the refined factor set $F^{*}$, the attribution instruction $Q$, and an exemplar critical sample $x_{\mathrm{ex}}$, and is then reused across critical samples from the same task. This design provides a stable and reusable reasoning structure for subsequent neuro-symbolic attribution. The generation and validation process is illustrated in Fig.~\ref{fig:subtask_graph} and summarized in Lines 4--9 of Algorithm~\ref{alg:nesy_csa}.

Specifically, given the refined key attribution factors $F^{*}$, the global attribution instruction $Q$, and a randomly selected exemplar critical sample $x_{\mathrm{ex}}$, NeSy-CSA prompts an LLM-based planner to generate a candidate subtask graph in a fixed JSON schema:
\begin{equation}
\widehat{G}=\{ST_{1},ST_{2},\dots,ST_{T}\}.
\end{equation}
Each node $ST_t$ denotes a subtask and is represented as
\begin{equation}
ST_t=\langle g_t,X_t,O_t,D_t\rangle,
\end{equation}
where $g_t$ specifies the objective of the $t$-th subtask, $X_t$ denotes its structured input interface, $O_t$ denotes its output handle, and $D_t$ denotes the set of  subtasks on which it depends.


To ensure that the generated graph supports valid stepwise attribution, NeSy-CSA applies a deterministic validation procedure before execution. The validation checks three aspects: \emph{continuity}, which requires legal and continuous step indices; \emph{predecessor dependency validity}, which ensures that each dependency refers only to an existing previous subtask; and \emph{input--output interface consistency}, which requires the output of each referenced predecessor to match the input expected by the dependent subtask.

Formally, the validity of the candidate graph is defined as
\begin{equation}
\mathrm{Valid}(\widehat{G})
=
\mathrm{Cont}(\widehat{G})
\wedge
\mathrm{Pred}(\widehat{G})
\wedge
\mathrm{IO}(\widehat{G}),
\end{equation}
where $\mathrm{Cont}(\widehat{G})$ checks continuity and legal step indexing, $\mathrm{Pred}(\widehat{G})$ checks whether each dependency refers only to an existing predecessor subtask, and $\mathrm{IO}(\widehat{G})$ checks whether the output interface of each referenced predecessor matches the required input interface of the dependent subtask.

In implementation, the predecessor set $D_t$ is recovered by parsing the output-handle references appearing in $X_t$. Any missing handle, out-of-range reference, non-predecessor dependency, or input--output mismatch is marked invalid and triggers regeneration with validator feedback. After validation, the resulting shared predecessor-dependent subtask graph is denoted by $G^{*}$.

Accordingly, this stage restricts the explanation space from $\mathcal{X}_{F}$ to
\begin{equation}
\mathcal{X}_{G}
=
\{h\in\mathcal{X}_{F}\mid h \text{ is derivable under } G^{*}\}.
\end{equation}
As a result, attribution is constrained to a validated and reusable predecessor-dependent graph, so that subsequent reasoning can proceed along an explicit evidence path rather than an implicit generation process.

\subsection{Neuro-Symbolic Hybrid Reasoning}

The third stage of NeSy-CSA executes the validated predecessor-dependent subtask graph $G^{*}$ through neuro-symbolic attribution. Open-ended critical scenario attribution cannot be fully handled by existing neuro-symbolic methods: although they improve flexibility by combining neural models with symbolic reasoning, their symbolic components often rely on predefined knowledge bases that are difficult to construct and cannot cover all tasks. Meanwhile, some attribution steps require semantic interpretation rather than complete formalization, whereas delegating the whole process to an unconstrained LLM may lead to unsupported conclusions. NeSy-CSA therefore routes each subtask according to its reasoning characteristics. Formalizable subtasks are executed through symbolic reasoning built from reusable atomic functions, avoiding the need for a comprehensive task-specific knowledge base, while semantic subtasks are handled by neural reasoning guided by predecessor evidence. This allows NeSy-CSA to combine interpretable reasoning for formalizable subtasks with evidence-grounded reasoning for semantic subtasks. The execution procedure is summarized in Lines 10--20 of Algorithm~\ref{alg:nesy_csa}. 

\paragraph{Neural Subtask Interpretation and Routing}

Given the validated graph $G^{*}$, NeSy-CSA first uses the neural component to interpret each subtask $ST_t$ and determine its execution mode. The key question is whether the subtask can be grounded as an explicit procedure whose result is checkable, or whether it requires semantic interpretation beyond fixed computation. Based on this interpretation, the execution mode of $ST_t$ is defined as
\begin{equation}
r_t=
\begin{cases}
\mathrm{SymExec}, &
\mathrm{Formalizable}(ST_t)\wedge \mathrm{Verifiable}(ST_t),\\
\mathrm{NeuInfer}, & \text{otherwise.}
\end{cases}
\label{eq:routing_new}
\end{equation}
Here, $\mathrm{Formalizable}(ST_t)$ indicates that the subtask can be expressed through deterministic computation, logical rules, or executable procedures, while $\mathrm{Verifiable}(ST_t)$ indicates that its output can be checked through re-execution, consistency tests, or explicit criteria. This neural interpretation acts as the routing interface between the subtask graph and the hybrid executor: subtasks that can be both formalized and verified are assigned to symbolic execution, whereas the remaining subtasks are handled by neural inference under dependency constraints.

\paragraph{Symbolic Execution}



When $r_t=\mathrm{SymExec}$, the subtask is assigned to the symbolic branch of NeSy-CSA. Instead of relying on a complete task-specific rule system, NeSy-CSA uses the LLM to dynamically construct the symbolic procedure required by each formalizable subtask, based on the semantics of the subtask and a lightweight atomic function library. The constructed procedure is then executed on the scenario data to obtain an explicit symbolic result, avoiding the need to predefine comprehensive attribution rules for each task.

Specifically, NeSy-CSA maintains a lightweight library of atomic functions:
\begin{equation}
\mathcal{A}=\{\omega_1,\omega_2,\dots,\omega_L\},
\end{equation}
where each atomic function $\omega_l$ implements a basic measurable operation, such as geometric computation, temporal comparison, logical predicate evaluation, or constraint checking. These functions do not encode complete task-specific attribution rules; instead, they serve as reusable primitives for constructing symbolic subtasks.

For each symbolic subtask $ST_t$, NeSy-CSA constructs an executable symbolic procedure $\Pi_t$. Depending on the required operation, $\Pi_t$ is instantiated either as a single atomic function or as a task-specific procedure constructed from the atomic function library:


\begin{equation}
\Pi_t =
\begin{cases}
\omega_l,
& ST_t \in \mathcal{B},\\[1mm]
\mathrm{Prog}_t(ST_t,\mathcal{A}),
& ST_t \in \mathcal{C},
\end{cases}
\end{equation}
where $\mathcal{B}$ denotes symbolic subtasks that can be directly handled by a single atomic function, and $\mathcal{C}$ denotes those requiring a task-specific executable procedure. Here, $\omega_l\in\mathcal{A}$, and $\Pi_t$ is the symbolic procedure constructed for $ST_t$. The procedure $\mathrm{Prog}_t(ST_t,\mathcal{A})$ is generated by the LLM according to the subtask semantics, with the verified atomic functions in $\mathcal{A}$ serving as executable building blocks. To ensure verifiability, the generated procedure is accepted only after passing execution and consistency checks.

The symbolic result is then obtained by executing the constructed procedure on the scenario and trajectory:

\begin{equation}
e_t^{\mathrm{sym}}
=
\Pi_t(s_i,\tau_i).
\end{equation}

This formulation makes the formalizable parts of attribution interpretable, while avoiding the need to predefine a complete high-level rule base for each task.

\paragraph{Neural Inference}


When $r_t=\mathrm{NeuInfer}$, the subtask is assigned to the neural branch of NeSy-CSA. This branch handles subtasks that cannot be reliably reduced to executable symbolic reasoning, such as those involving incomplete evidence, contextual interpretation, or comparison among multiple plausible causes. NeSy-CSA employs an LLM to perform neural inference for semantic subtasks, while constraining its input with the predecessor evidence:

\begin{equation}
e_t^{\mathrm{neu}}
=
\mathrm{NeuInfer}\!\left(
ST_t,s_i,\tau_i,\{e_j\mid j\in D_t\},K_{\mathrm{dom}}
\right).
\end{equation}
Here, $\mathrm{NeuInfer}(\cdot)$ denotes the LLM-based neural inference operator, $D_t$ is the predecessor set of $ST_t$, and $\{e_j\mid j\in D_t\}$ provides the intermediate results produced by its valid predecessor subtasks. By using these dependency-constrained results as part of the neural input, NeSy-CSA limits the LLM to evidence relevant to the current subtask. This preserves the semantic flexibility of LLMs for open-ended attribution while reducing unsupported or inconsistent reasoning.

With the two execution branches defined above, NeSy-CSA executes the validated graph $G^{*}$ in dependency order. For sample $x_i$, each subtask result is produced either by symbolic execution or by  neural inference, yielding the complete result sequence
\begin{equation}
E_i=\{e_t\}_{t=1}^{T},
\end{equation}
where each $e_t$ is produced either by symbolic execution for formalizable and verifiable subtasks or by dependency-constrained neural inference for the remaining subtasks under the routing rule in \eqref{eq:routing_new}. Accordingly, the execution stage further restricts the explanation space from $\mathcal{X}_{G}$ to
\begin{equation}
\mathcal{X}_{R}
=
\{h\in\mathcal{X}_{G}\mid h \text{ is constructible from } E_i\}.
\end{equation}
Finally, the result of the terminal subtask is taken as the attribution conclusion for sample $x_i$:
\begin{equation}
R_i^{\star}:=e_T,\qquad R_i^{\star}\in\mathcal{X}_{R}.
\end{equation}

\subsection{Evaluation of Open-Ended Attribution}


Evaluating critical scenario attribution is challenging because open-ended attribution cannot be reduced to selecting from a fixed set of standard answers. NeSy-CSA therefore uses a two-level evaluation: a process-level assessment examines the structural validity of the generated reasoning workflow, while a result-level assessment evaluates the behavioral effectiveness of the attribution results.

\paragraph{Process-level Assessment}

Since open-ended attribution does not admit a unique gold-standard reasoning chain, we evaluate the generated reasoning workflow by measuring its semantic consistency with environment-specific reference procedures. Let the generated subtask graph be denoted by \(G^*=\langle ST_1,\ldots,ST_T\rangle\). For each target environment, three researchers familiar with the environment independently define reference subtask sets before the experiments. The reference set provided by the \(r\)-th researcher is denoted as \(L^{(r)}=\{l^{(r)}_1,\ldots,l^{(r)}_{M_r}\}\), where \(r=1,2,3\) and \(M_r\) is the number of reference subtasks in the \(r\)-th set. These reference sets are fixed before evaluation and are used for all evaluated graphs without post-hoc adjustment. For the \(r\)-th expert-defined reference set, semantic similarity is computed using a text encoder \(\phi(\cdot)\):
\begin{equation}
s^{(r)}(ST_i,l^{(r)}_j)
=
\cos\!\big(\phi(ST_i),\phi(l^{(r)}_j)\big).
\end{equation}

For each generated subtask \(ST_i\), we identify its best-matched reference step in the \(r\)-th reference set by
\begin{equation}
j_r^{*}(i)
=
\arg\max_{1\le j\le M_r} s^{(r)}(ST_i,l^{(r)}_j).
\end{equation}

To evaluate both subtask validity and reference coverage, we compute process-level Precision, Recall, and F1 with respect to each expert-defined reference set. Let \(TP^{(r)}\) denote the set of distinct reference subtasks covered by the generated reasoning workflow under the \(r\)-th reference set:
\begin{equation}
TP^{(r)}
=
\left\{
j_r^*(i)
\mid
s^{(r)}\!\left(ST_i,l^{(r)}_{j_r^*(i)}\right)\ge \eta_m
\right\}.
\end{equation}

Then Precision, Recall, and the F1 score with respect to the \(r\)-th reference set are defined as
\begin{equation}
\begin{gathered}
\mathrm{Precision}^{(r)}
=
\frac{|TP^{(r)}|}{T},
\qquad
\mathrm{Recall}^{(r)}
=
\frac{|TP^{(r)}|}{M_r},\\
\mathrm{F1}^{(r)}
=
\frac{
2\cdot \mathrm{Precision}^{(r)}\cdot \mathrm{Recall}^{(r)}
}{
\mathrm{Precision}^{(r)}+\mathrm{Recall}^{(r)}
}.
\end{gathered}
\end{equation}

Since open-ended attribution may admit multiple valid reasoning procedures, the generated subtask graph is evaluated against all three expert-defined reference sets. The best-matched reference set is selected according to the F1 score:
\begin{equation}
r^*
=
\arg\max_{r\in\{1,2,3\}}
\mathrm{F1}^{(r)}.
\end{equation}

The final process-level scores are reported as
\begin{equation}
\begin{gathered}
\mathrm{Precision}
=
\mathrm{Precision}^{(r^*)},
\qquad
\mathrm{Recall}
=
\mathrm{Recall}^{(r^*)},\\
\mathrm{F1}
=
\mathrm{F1}^{(r^*)}.
\end{gathered}
\end{equation}
Here, \(j_r^{*}(i)\) denotes the best-matched reference subtask for the generated subtask \(ST_i\) in the \(r\)-th reference set, \(s^{(r)}(\cdot,\cdot)\) is the semantic similarity score, and \(\eta_m\) is the matching threshold. \(T\) and \(M_r\) denote the numbers of generated subtasks and reference subtasks in the \(r\)-th reference set, respectively. The set \(TP^{(r)}\) removes duplicate matches, so each covered reference subtask is counted only once.





\paragraph{Result-level Assessment}

\begin{figure}[t]
    \centering
    \includegraphics[width=\linewidth]{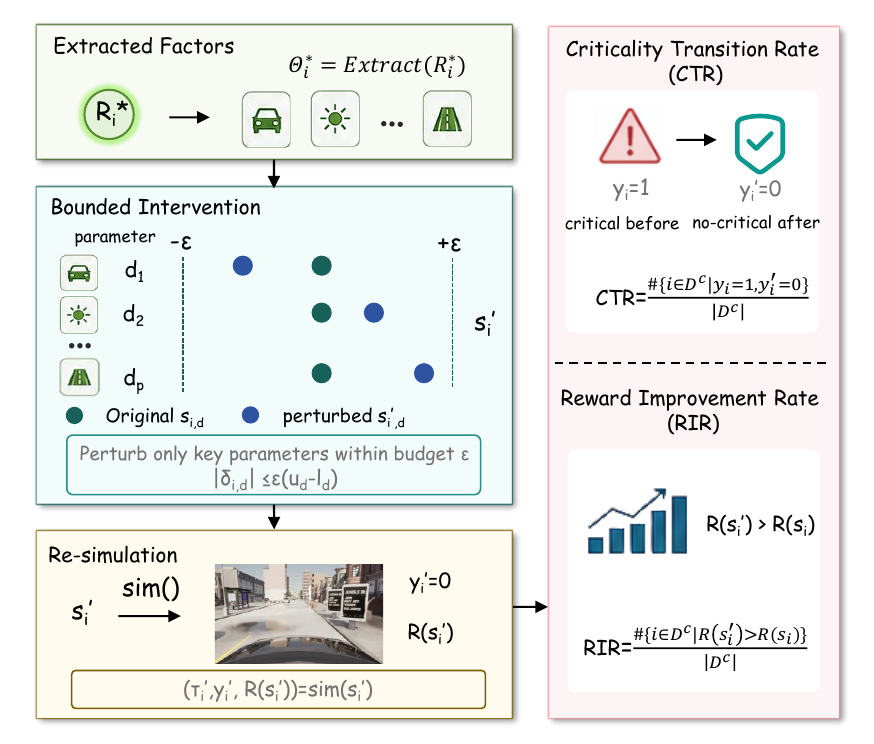}
    \caption{
   Result-level assessment procedure based on bounded intervention, re-simulation, and CTR/RIR computation. 
    }
    \label{fig:result_level_assessment}
    \vspace{-0.5cm}
\end{figure}

Process-level alignment only indicates that the generated reasoning procedure is structurally reasonable; it does not show whether the attribution conclusion is effective. Since critical scenario attribution aims to identify the contributors to criticality, we further evaluate whether modifying the relevant variables extracted from the attribution results can improve the scenario outcome under a bounded perturbation budget. This allows the behavioral effect of the attribution result to be tested through simulation. Fig.~\ref{fig:result_level_assessment} provides an overview of the result-level assessment.

Given an attribution conclusion $R_i^{\star}$ for a critical sample $x_i$, the scenario parameters identified by the attribution result are extracted as an index set
\begin{equation}
\Theta_i^{*}=\mathrm{Extract}(R_i^{\star}).
\end{equation}

A perturbed scenario is then constructed by modifying only the extracted parameters according to the direction suggested by the attribution result:

\begin{equation}
\begin{aligned}
s_i' &= s_i + \Delta_i, \\
\Delta_{i,d} &=
\begin{cases}
0, & d\notin\Theta_i^{*},\\[2pt]
\rho_{i,d}\,\varepsilon_{i,d}(u_d-l_d), & d\in\Theta_i^{*},
\end{cases}
\end{aligned}
\end{equation}
where $[l_d,u_d]$ denotes the valid range of the $d$-th parameter, $\varepsilon_{i,d}$ bounds the perturbation magnitude ($0<\varepsilon_{i,d}\le \varepsilon$), and $\rho_{i,d}\in\{-1,+1\}$ indicates the modification direction inferred from the attribution result. For example, if excessive speed is identified as a contributor to collision, the corresponding speed parameter is reduced within the bounded range before re-simulation.

Based on the critical set $\mathcal{D}^{c}=\{i\mid y_i=1\}$, we define two complementary metrics computed from the same perturbation:

\begin{equation}
\begin{aligned}
\mathrm{CTR}
&=
\frac{1}{|\mathcal{D}^{c}|}
\sum_{i\in \mathcal{D}^{c}}
\mathbf{1}\!\left[y_i'=0\right],\\
\mathrm{RIR}
&=
\frac{1}{|\mathcal{D}^{c}|}
\sum_{i\in \mathcal{D}^{c}}
\mathbf{1}\!\left[R(s_i')>R(s_i)\right].
\end{aligned}
\end{equation}

CTR measures how often the perturbation on the identified parameters turns critical samples into non-critical ones, while RIR measures how often it increases the episode reward. Together, they provide direct behavioral evidence for the effectiveness of the attribution result: if the extracted parameters are truly relevant to criticality, modifying them should change the criticality outcome or improve the reward.

\section{Experiments}


This section evaluates NeSy-CSA in terms of attribution effectiveness, process quality, and component contribution. We first describe the experimental environments, parameter settings, and baselines. We then evaluate attribution effectiveness across diverse decision-making tasks and assess the quality and efficiency of the predecessor-dependent subtask graph. Next, ablation studies are conducted to analyze the contribution of key components. Finally, we use a case study to demonstrate the attribution process, providing a direct view of NeSy-CSA’s traceable and interpretable reasoning.

\subsection{Experimental Setup}

\subsubsection{Environments}

\begin{figure}[t]
  \centering
  \includegraphics[width=0.9\columnwidth]{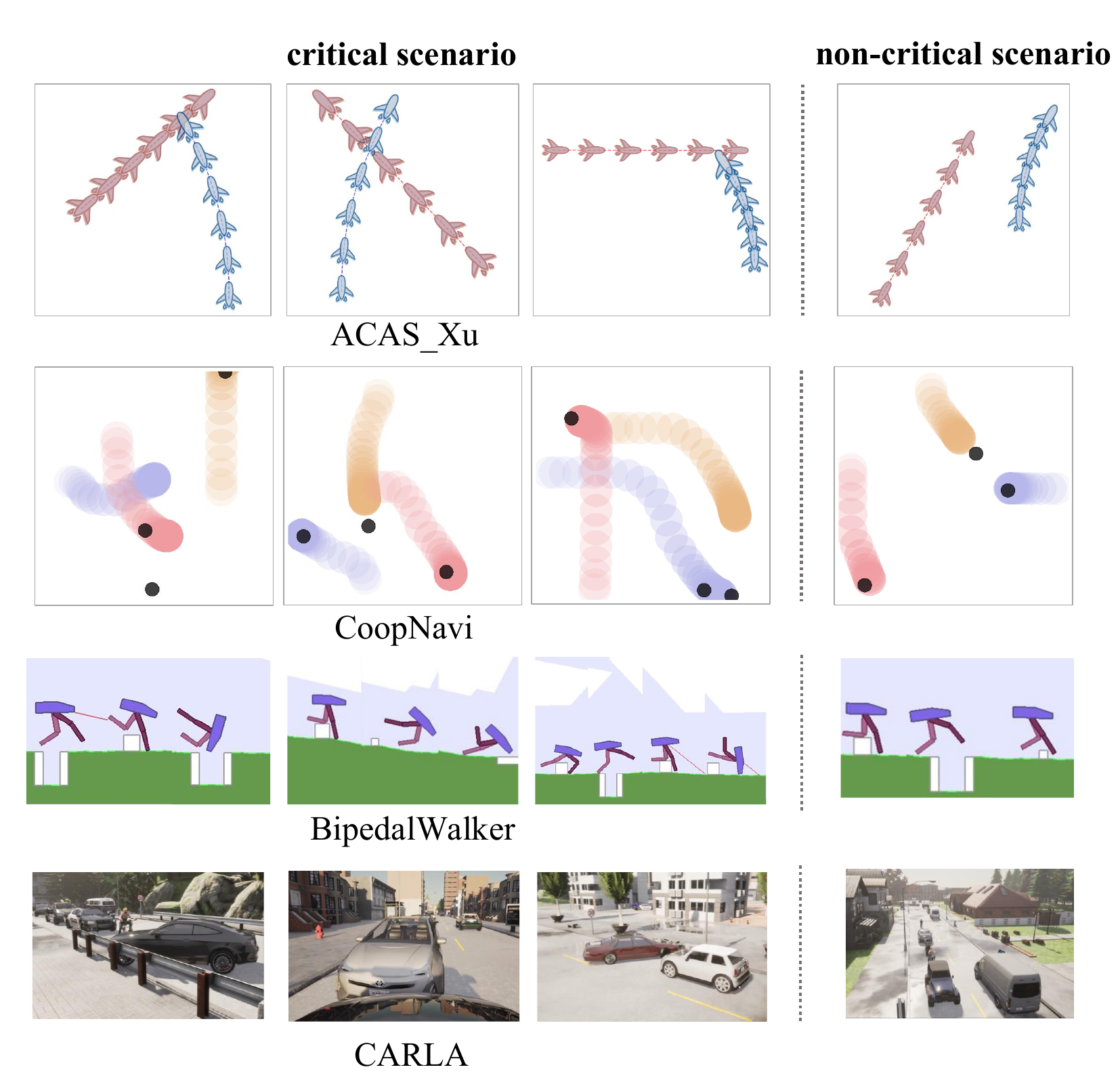}
  \caption{\textbf{Illustrative critical scenarios across four environments.}
ACAS\_Xu: unsafe separation between the ownship (blue) and the intruder (red).
CoopNavi: incomplete team coordination in reaching assigned landmarks.
BipedalWalker: loss of locomotion stability on diverse terrains.
CARLA: urban driving scenarios involving collision or unsuccessful destination arrival.}
  \label{fig:envs}
\vspace{-0.4cm}
  
\end{figure}

We evaluate NeSy-CSA on four decision-making environments with different state spaces, interactions, and failure mechanisms, as illustrated in Fig.~\ref{fig:envs}. 

\textbf{ACAS\_Xu:} Airborne collision avoidance for drones and small aircraft. Critical scenarios occur when advisories fail to maintain safe separation.

\textbf{CoopNavi:} Cooperative multi-agent navigation. Criticality arises when poor coordination, target conflicts, or collision-prone interactions cause agents to exceed the collision limit or fail to reach their assigned landmarks.

\textbf{BipedalWalker:} Bipedal locomotion on 2D terrain. Critical scenarios happen when body states, contact patterns, or terrain conditions lead to falling or loss of progress.

\textbf{CARLA:} Urban autonomous driving. Critical scenarios occur when the ego vehicle cannot handle road geometry, traffic participants, or route-following constraints, leading to collisions or task time-outs.

\subsubsection{Experimental Data}

Table~\ref{tab:dataset_stats} summarizes the dataset statistics for each environment. 
For ACAS\_Xu, CoopNavi, BipedalWalker, and CARLA, we report the total number of scenarios $|\mathcal{D}|$, as well as the number of critical ($|\mathcal{D}^{c}|$) and non-critical ($|\mathcal{D}^{nc}|$) samples.

\begin{table}[t]
\centering
\caption{Dataset statistics across environments.}
\label{tab:dataset_stats}
\renewcommand{\arraystretch}{1.08}
\setlength{\tabcolsep}{3.2pt}
\footnotesize
\begin{tabular*}{\columnwidth}{@{\extracolsep{\fill}}lcccc}
\toprule
Metric & ACAS\_Xu & CoopNavi & BipedalWalker & CARLA \\
\midrule
Total $|\mathcal{D}|$          & 6,000 & 2,000 & 2,000 & 3,000 \\
Non-critical $|\mathcal{D}^{nc}|$   & 5,799 & 1,789 & 1,764 & 2,800 \\
Critical $|\mathcal{D}^{c}|$ & 201   & 211   & 236   & 200 \\
\bottomrule
\vspace{-0.75cm}
\end{tabular*}
\end{table}

All scenarios are generated using the MDPFuzz procedure, producing critical and non-critical cases under varied task conditions. 
The full datasets are used for key attribution factor refinement by comparing critical and non-critical samples. 
For result-level evaluation, 200 critical scenarios are selected from each environment to assess whether interventions on the attributed factors can mitigate scenario criticality. 
This design ensures consistent evaluation across environments and focuses analysis on long-tail critical events, which are the main target in scenario-based testing.

\subsubsection{Parameter Settings}

NeSy-CSA uses four main parameters: the statistical significance threshold $\alpha$, the effect-size threshold $\delta$, the semantic similarity thresholds $\eta_d$ and $\eta_m$, and the intervention budget $\varepsilon$. These parameters are fixed for all main experiments.

\textbf{Statistical significance threshold $\alpha$.}
In the key attribution factor refinement stage, we set $\alpha=0.05$ for screening candidate factors. This is a standard significance level for testing whether a candidate factor differs between critical and non-critical samples. Since multiple factors are tested simultaneously, the Benjamini--Hochberg procedure is applied to control false discoveries.

\textbf{Effect-size threshold $\delta$.}
We set $\delta=0.5$, following the conventional benchmark for a medium effect size~\cite{cohen1988statistical}. This threshold is used to remove factors whose differences are statistically significant but practically weak, so that the retained factors are more informative for subsequent attribution.

\textbf{Semantic similarity thresholds $\eta_d$ and $\eta_m$.}
Both thresholds are used to determine sentence-level semantic similarity: $\eta_d$ for deduplicating candidate attribution factors, and $\eta_m$ for comparing generated subtasks with reference steps in process-level evaluation. We set $\eta_d=\eta_m=0.7$ in the experiments, providing a practical balance between accepting reasonable paraphrases and rejecting loose matches.

\begin{figure}[t]
    \centering
    \includegraphics[width=0.85\columnwidth]{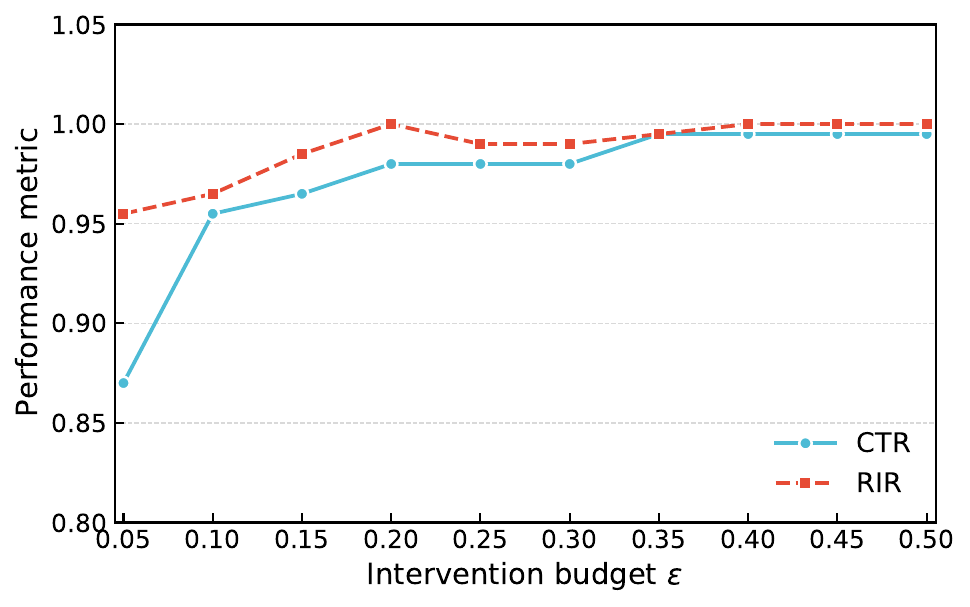}
    \caption{Sensitivity analysis of the intervention budget $\varepsilon$ on ACAS\_Xu. }
    \label{fig:epsilon_sensitivity}
\vspace{-0.6cm}
\end{figure}

\textbf{Intervention budget $\varepsilon$.} In result-level evaluation, only the scenario parameters extracted from the attribution result are perturbed, and the perturbation magnitude is bounded by the valid range of each parameter. The intervention budget $\varepsilon$ controls the maximum perturbation scale. We determine this value through environment-wise parameter calibration. Specifically, we sweep different values of $\varepsilon$ in each environment and select the value that reaches a stable improvement region while avoiding unnecessarily large deviations from the original critical scenario. Fig.~\ref{fig:epsilon_sensitivity} shows the calibration result on ACAS Xu as an example. CTR and RIR increase rapidly when $\varepsilon$ is small and become stable around $\varepsilon=0.3$, after which larger perturbations bring only marginal improvement while reducing locality. Therefore, we set $\varepsilon=0.3$ as the intervention budget. 


\subsubsection{Baseline Settings}

We compare NeSy-CSA with three LLM-based baselines to analyze the contribution of different reasoning configurations in critical scenario attribution.




\textbf{LLM-only:} The LLM directly performs attribution in an end-to-end manner by prompting with scenario data, trajectory information, and task instructions, producing final attribution conclusions without access to external tools or structured intermediate representations.


\textbf{LLM+Tool:} The LLM augments its reasoning process with a predefined library of atomic tools $\mathcal{A}$, where each tool implements an elementary and executable function, including geometric computation, temporal comparison, logical predicate evaluation, and constraint checking. Given scenario data, trajectory information, and task instructions, the LLM performs single-pass inference to generate a reasoning response while optionally invoking tools when intermediate computation or verification is required. Tool invocation is performed by directly mapping the current reasoning context to a tool call, and each tool is executed independently without forming multi-step tool chains. The final attribution result is obtained by integrating tool outputs within the same generation process.

\textbf{LLM+CoT:} The LLM performs attribution using structured step-by-step reasoning prompting within a single context window. The LLM is guided to decompose the attribution process into a sequence of intermediate reasoning steps, including: (i) extraction of scenario evidence from initial states and trajectory observations, (ii) identification of failure phenomena exhibited in the scenario, (iii) causal reasoning that links observed failure behaviors to relevant state variables, and (iv) synthesis of the final attribution conclusion. The reasoning process is conducted entirely in natural language without access to external tools or structured symbolic execution mechanisms.

\subsubsection{Implementation Details}
NeSy-CSA uses DeepSeek-V3.2 for semantic subtasks (e.g., candidate attribution-factor generation and neural inference) and routes multimodal subtasks to GPT-4o-mini to balance reasoning quality and cost.



\begin{table*}[t]
\centering
\caption{Attribution effectiveness comparison with LLM-based baselines.}
\label{tab:main_results}
\renewcommand{\arraystretch}{1.15}
\setlength{\tabcolsep}{5pt}
\begin{tabular*}{\textwidth}{@{\extracolsep{\fill}}ccccccccc@{}}
\toprule
\multirow{2}{*}{Method}
& \multicolumn{2}{c}{ACAS Xu}
& \multicolumn{2}{c}{CoopNavi}
& \multicolumn{2}{c}{BipedalWalker}
& \multicolumn{2}{c}{CARLA} \\
\cmidrule(lr){2-3}
\cmidrule(lr){4-5}
\cmidrule(lr){6-7}
\cmidrule(lr){8-9}
& CTR (\%) & RIR (\%) 
& CTR (\%) & RIR (\%) 
& CTR (\%) & RIR (\%) 
& CTR (\%) & RIR (\%) \\
\midrule
LLM-only 
& 71.33±3.01 & 86.33±0.29 
& 59.33±2.26 & 62.67±2.36 
& 79.50±3.61 & 86.33±3.01 
& 47.50±0.50 & 75.33±1.61 \\

LLM+Tool 
& 62.67$\pm$2.25 & 82.33$\pm$4.51 
& 51.83±3.06 & 56.83±4.25 
& 80.17±0.76 & 91.17±0.76 
& 50.00±2.16 & 77.00±1.78 \\

LLM+CoT  
& 59.83$\pm$1.76 & 81.67$\pm$2.52 
& 44.00±12.85 & 50.33±10.97 
& 70.67±0.24 & 81.83±0.62 
& 53.83±7.18 & 75.67±2.08 \\

NeSy-CSA 
& \textbf{99.50±0.50} & \textbf{99.83±0.29} 
& \textbf{69.33±1.26} & \textbf{84.50±3.50} 
& \textbf{87.00±1.78} & \textbf{93.50±0.82} 
& \textbf{61.00±2.00} & \textbf{79.33±2.52} \\
\bottomrule
\vspace{-0.9cm}
\end{tabular*}
\end{table*}

\subsection{Attribution Effectiveness Compared with Baselines}

We evaluate the result-level attribution effectiveness of NeSy-CSA against three LLM-based baselines, including LLM-only, LLM+Tool, and LLM+CoT. The comparison is conducted on four environments using CTR and RIR, with each experiment repeated three times and averaged.

Table~\ref{tab:main_results} reports the result-level comparison. NeSy-CSA achieves the highest CTR and RIR in all four environments. Averaged over the four environments, it improves CTR by 18.32\% and RIR by 13.67\% compared with the mean performance of the three LLM-based baselines. These gains indicate that NeSy-CSA produces more accurate attribution conclusions, as the parameters extracted from these conclusions are more likely to change the critical outcome under targeted perturbations. The largest CTR improvement appears in ACAS Xu, where the criticality pattern is relatively concentrated and largely determined by the relative geometry and motion trends between aircraft. Therefore, once the attribution conclusion correctly captures these relations, the extracted intervention parameters can more directly affect the separation condition, leading to larger CTR improvements. A substantial gain is also observed in CoopNavi. Although this environment involves more complex multi-agent coordination, many critical cases are still closely related to spatial configurations and interaction patterns; accurate attribution can therefore guide changes to agent positions or relational layouts that improve the outcome. The improvements in BipedalWalker and CARLA are more moderate. In BipedalWalker, unsuccessful rollouts often involve coupled contact and motion dynamics, so changing a few extracted parameters may only partially affect the trajectory. In CARLA, the RIR gain is especially limited because some critical cases correspond to highly constrained traffic or route conditions, where bounded perturbations of the extracted parameters may not be sufficient to recover the task. Overall, these results demonstrate that NeSy-CSA provides more accurate and effective attribution across different types of decision-making scenarios.

The results also show that simply adding tools or step-by-step reasoning does not consistently improve attribution effectiveness. In ACAS Xu, LLM+Tool and LLM+CoT are lower than LLM-only by 8.66\% and 11.50\% in CTR, and by 4.00\% and 4.66\% in RIR, respectively. In CoopNavi, LLM+CoT is lower than LLM-only by 15.33\% in CTR and 12.34\% in RIR. Although LLM+Tool improves RIR in BipedalWalker and LLM+CoT achieves the best baseline CTR in CARLA, no LLM-based baseline consistently improves both CTR and RIR across environments. This suggests that the key limitation is not whether extra computation or explicit reasoning steps are available, but whether the open-ended attribution space is progressively constrained before the final conclusion is generated. Specifically, LLM+Tool introduces numerical computations, but it does not refine the attribution factor space or compare candidate evidence between critical and non-critical groups. As a result, many measurable but non-key variables may enter the reasoning process and shift the attribution toward irrelevant factors. Moreover, the predefined tool set only provides partial local measurements, rather than all precise computations required by task-specific attribution. LLM+CoT exposes intermediate reasoning steps, but these steps are still generated by the LLM without validated dependencies or executable checks. Therefore, unverified or hallucinated intermediate statements may accumulate and lead to worse attribution results in some environments.

NeSy-CSA addresses this bottleneck by constraining the attribution process before the final conclusion is produced. Factor refinement reduces the open-ended candidate space using both semantic relevance and data-driven differences between critical and non-critical samples. The predecessor-dependent subtask graph organizes the selected evidence along validated dependencies, so that later judgments are not based on isolated observations. Neuro-symbolic execution further checks formalizable relations through executable symbolic procedures and constrains semantic inference with predecessor evidence. These constraints reduce the influence of irrelevant evidence and unsupported intermediate judgments, making the final attribution conclusion more reliable and better grounded in the observed critical behavior.



\subsection{Process-Level Quality and Efficiency of Subtask Decomposition}

\begin{table}[t]
\centering
\caption{Process-level quality of the generated predecessor-dependent subtask graphs.}
\label{tab:process_quality}
\renewcommand{\arraystretch}{1.08}
\setlength{\tabcolsep}{3.2pt}
\footnotesize
\begin{tabular*}{\columnwidth}{@{\extracolsep{\fill}}cccc}
\toprule
Environment & Precision (\%) & Recall (\%) & F1 (\%) \\
\midrule
ACAS\_Xu       & 100.00±0.00 & 100.00±0.00 & 100.00±0.00 \\
CoopNavi       & 83.33±14.43 & 83.07±4.58  & 82.50±4.33 \\
BipedalWalker  & 77.78±9.62  & 84.13±1.37  & 80.56±4.81  \\
CARLA          & 83.07±15.00 & 86.77±14.40  & 84.35±11.67 \\
\bottomrule
\vspace{-0.7cm}
\end{tabular*}
\end{table}

\begin{table}[t]
\centering
\caption{Comparison between per-sample and one-time subtask graph decomposition.}
\label{tab:process_quality_efficiency}
\renewcommand{\arraystretch}{1.08}
\setlength{\tabcolsep}{3.2pt}
\footnotesize
\begin{tabular}{ccccc}
\toprule
Environment & Strategy & CTR (\%) & RIR (\%) & Tokens ($\times 10^3$) \\
\midrule
\multirow{2}{*}{ACAS\_Xu}
& Per-sample & 99.50±0.50& 99.67±0.29& 2234.97±21.05\\
& Ours       & 99.50±0.50& 99.83±0.29& 10.27±0.01\\
\midrule
\multirow{2}{*}{CoopNavi}& Per-sample & 58.33±3.21& 78.50±3.12& 986.18±12.92\\
& Ours       & 69.33±1.26& 84.50±3.50& 3.98±0.20\\
\midrule
\multirow{2}{*}{BipedalWalker}& Per-sample & 82.50±1.80& 91.33±3.10& 747.42±2.40\\
& Ours       & 87.00±1.78& 93.50±0.82& 3.55±0.36\\
\midrule
\multirow{2}{*}{CARLA}& Per-sample & 56.17±3.06& 77.33±2.75& 1093.03±70.87\\
& Ours       & 61.00±2.00& 79.33±2.52& 4.19±0.35\\
\bottomrule
\end{tabular}

\vspace{2pt}
\footnotesize{Token cost is reported for the decomposition stage only.}
\vspace{-0.6cm}
\end{table}

Besides the result-level evaluation based on CTR and RIR, we further assess the process-level quality of NeSy-CSA by examining whether the predecessor-dependent subtask graph provides a valid reasoning structure for critical scenario attribution. Since the LLM-based baselines do not explicitly construct such a graph, this evaluation focuses on the internal reasoning process of NeSy-CSA rather than serving as another baseline comparison.

Table~\ref{tab:process_quality} reports the process-level Precision, Recall, and F1 of the generated subtask graphs. For each environment, the generated graph is evaluated against three independently defined expert reference procedures, and the best-matched reference procedure is selected according to the F1 score. The reported scores are averaged over three runs. Overall, the generated graphs show high consistency with expert procedures, achieving average Precision, Recall, and F1 of 86.05\%, 88.49\%, and 86.85\%, respectively. This indicates that NeSy-CSA can construct a reasonable intermediate reasoning structure, rather than producing attribution results through an unconstrained generation process. Moreover, Recall is generally slightly higher than Precision, especially in BipedalWalker and CARLA, suggesting that the generated graphs tend to cover the key expert-defined attribution steps, while occasionally introducing extra or boundary-related subtasks in more complex environments. Thus, the main process-level issue is not missing critical reasoning steps, but limited redundancy caused by complex scenario dynamics.

Table~\ref{tab:process_quality_efficiency} compares the one-time decomposition with per-sample subtask decomposition. Across four environments, one-time decomposition achieves comparable or better attribution quality, with average CTR improving by 5.08\% and average RIR improving by 2.58\%, and no environment showing performance degradation. Meanwhile, one-time decomposition greatly reduces the decomposition-stage token cost, from $(1265.40\pm26.81)\times10^{3}$ to $(5.50\pm0.23)\times10^{3}$ tokens on average, corresponding to over two orders of magnitude reduction. In ACAS\_Xu, the reduction reaches 99.5\%. These results confirm that the high-level attribution structure is reusable within the same environment. Per-sample decomposition introduces redundant computations and over-specialized subtasks, while one-time decomposition constructs an environment-level reasoning model that maintains accuracy, suppresses sample-level noise, and eliminates per-sample overhead.

\subsection{Ablation Study on Key Components}

To evaluate the contribution of different components to attribution performance, we design the following ablation variants and report CTR and RIR.

\textbf{w/o Attribution Factor:} The entire factor refinement stage is removed. The method no longer constructs the refined factor set $F^{*}$ from critical and non-critical samples, so subsequent graph construction and attribution reasoning are driven directly by the task instruction and scenario evidence.

\textbf{w/o Subtask Graph:} Attribution is conducted without the predecessor-dependent subtask graph, so the model generates the final attribution result directly rather than through predecessor-dependent intermediate subtasks.

\textbf{w/o Symbolic Execution:} The subtask graph is retained, but formalizable subtasks are all handled by neural inference.

\begin{table*}[t]
\centering
\caption{Ablation study on key components of NeSy-CSA across four environments.}
\label{tab:ablation_components}
\renewcommand{\arraystretch}{1.15}
\setlength{\tabcolsep}{0pt}
\begin{tabular*}{\textwidth}{@{\extracolsep{\fill}}ccccccccccc@{}}
\toprule
\multirow{2}{*}{
\begin{tabular}[c]{@{}c@{}}Attribution\\Factor\end{tabular}}
&
\multirow{2}{*}{
\begin{tabular}[c]{@{}c@{}}Subtask\\Graph\end{tabular}}
&
\multirow{2}{*}{
\begin{tabular}[c]{@{}c@{}}Symbolic\\Execution\end{tabular}}
& \multicolumn{2}{c}{ACAS Xu}
& \multicolumn{2}{c}{CoopNavi}
& \multicolumn{2}{c}{BipedalWalker}
& \multicolumn{2}{c}{CARLA} \\
\cmidrule(lr){4-5}
\cmidrule(lr){6-7}
\cmidrule(lr){8-9}
\cmidrule(lr){10-11}
& & & CTR (\%) & RIR (\%) & CTR (\%) & RIR (\%) & CTR (\%) & RIR (\%) & CTR (\%) & RIR (\%) \\
\midrule
                  & \checkmark & \checkmark & 91.83±4.86 & 94.50±3.12 & 56.33±5.01 & 75.00±2.29 & 78.17±6.83  & 85.50±5.57 & 54.33±2.52 & 75.17±2.36 \\
\checkmark        &            & \checkmark & 82.00±1.32 & 88.83±1.44 & 60.33±5.03 & 64.17±3.45 & 79.33±1.15 & 89.50±1.50 & 55.25±2.72 & 77.25±2.63 \\
\checkmark        & \checkmark &            & 71.33±13.14 & 84.50±5.63 & 55.97±3.54 & 80.13±0.55 & 79.17±1.26 & 88.83±0.76 & 52.00±2.18 & 77.33±0.76 \\
\checkmark        & \checkmark & \checkmark & \textbf{99.50±0.50} & \textbf{99.83±0.29} & \textbf{69.33±1.26} & \textbf{84.50±3.50} & \textbf{87.00±1.78} & \textbf{93.50±0.82} & \textbf{61.00±2.00} & \textbf{79.33±2.52} \\
\bottomrule
\vspace{-0.6cm}

\end{tabular*}
\end{table*}

Table~\ref{tab:ablation_components} reports the ablation results of the three key components. The full NeSy-CSA achieves the best CTR and RIR across all four environments, indicating that all three components are necessary for accurate critical scenario attribution. Removing symbolic execution leads to the largest average CTR drop, from 79.21\% to 64.62\%, showing that executable verification is critical for identifying factors that can actually change the critical outcome. Removing the subtask graph causes the largest average RIR drop, from 89.29\% to 79.94\%, especially in CoopNavi, suggesting that structured reasoning is necessary for organizing interaction-related evidence. Removing attribution factor refinement also reduces the average CTR to 70.17\% and the average RIR to 82.54\%, indicating that constraining the candidate factor space is needed before reasoning and verification. Overall, the ablation results show that NeSy-CSA gains its attribution accuracy from the combination of factor-space constraint, process-level organization, and symbolic verification.

\begin{table}[t]
\centering
\caption{Effect of data-driven filtering on attribution performance and token cost.}
\label{tab:factor_filtering_cost}
\renewcommand{\arraystretch}{1.08}
\setlength{\tabcolsep}{3.2pt}
\footnotesize
\begin{tabular}{cccccc}
\toprule
Environment & Strategy & \#F & Tokens ($\times 10^5$) & CTR (\%) & RIR (\%) \\
\midrule
\multirow{2}{*}{ACAS Xu} 
& w/o Filter & 12 & 25.92±1.25 & 98.17±1.15 & 98.50±1.32 \\
& Ours      & 3  & 9.31±0.04  & 99.50±0.50& 99.83±0.29\\
\midrule
\multirow{2}{*}{CoopNavi} 
& w/o Filter & 10 & 81.23±1.97 & 63.67±3.01& 80.17±1.44\\
& Ours      & 2  & 65.91±2.43 & 69.33±1.26& 84.50±3.50\\
\midrule
\multirow{2}{*}{BipedalWalker} 
& w/o Filter & 10& 25.33±1.73& 81.83±2.02& 89.17±1.80\\
& Ours      & 3& 15.33±1.51& 87.00±1.78& 93.50±0.82\\
\midrule
\multirow{2}{*}{CARLA} 
& w/o Filter & 13 & 33.77±0.004 & 57.50±2.09& 77.33±0.58\\
& Ours      & 3  & 17.73±1.61  & 61.00±2.00 & 79.33±2.52 \\
\bottomrule
\end{tabular}

\vspace{2pt}
\footnotesize{Token cost is reported for the complete attribution process.}
\vspace{-0.5cm}

\end{table}

The above ablation removes the entire key attribution factor refinement module. We further isolate the data-driven filtering step within this module to examine whether retaining fewer selected factors can preserve attribution performance while reducing reasoning cost. The w/o Filter setting provides the LLM with all generated candidate factors, which brings more information but also expands the attribution space and introduces redundant or weakly relevant factors. In contrast, Ours keeps only a compact set of data-selected factors. As shown in Table~\ref{tab:factor_filtering_cost}, Ours reduces the average number of factors from 11.25 to 2.75, while achieving comparable or slightly better CTR and RIR. The performance gap is moderate because w/o Filter still provides statistical summaries of the non-critical and critical groups, allowing the LLM to partially identify unimportant factors during reasoning. However, this happens after all candidates have entered the prompt, resulting in higher token cost and a noisier reasoning process. Therefore, data-driven filtering is valuable because it removes weak or redundant factors before attribution reasoning, enabling NeSy-CSA to obtain similar or better attribution performance with fewer and more informative factors.

\subsection{Case Study of Traceable Attribution Reasoning}

\begin{figure*}[t]
  \vspace{-0.2cm}
  \centering
  \includegraphics[width=\textwidth]{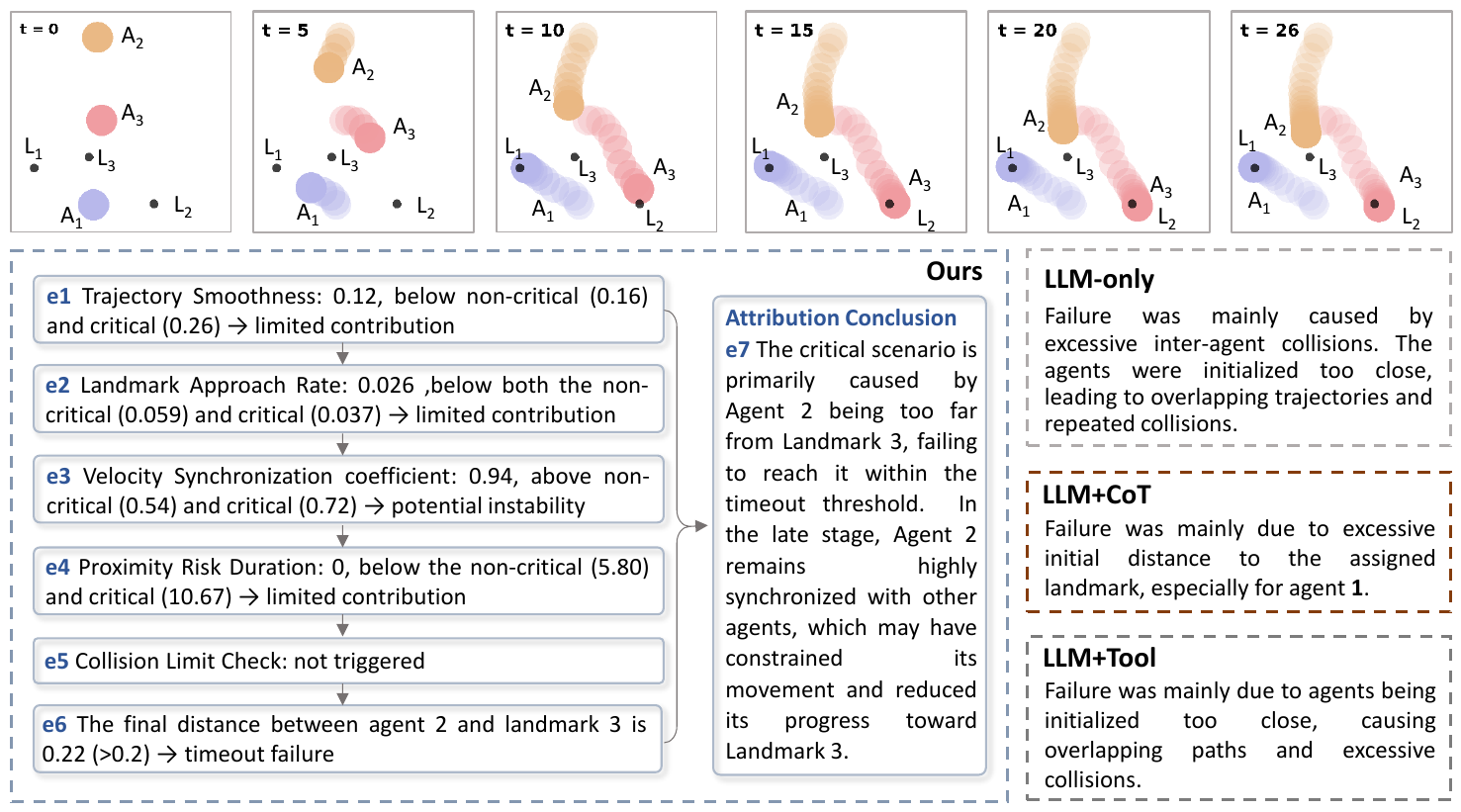}
  \caption{Case study of traceable neuro-symbolic attribution reasoning. }
  \label{fig:case_study}
\vspace{-0.5cm}
\end{figure*}

The previous quantitative experiments have evaluated result-level attribution effectiveness and process-level graph quality across different environments. This section further visualizes how NeSy-CSA and LLM-based baselines produce attribution results. Different from the process-level F1 evaluation, which measures the coverage of expert-defined subtask procedures, this case study illustrates the concrete attribution trace of NeSy-CSA, showing how intermediate evidence is used to support the final attribution conclusion.

As shown in Fig.~\ref{fig:case_study}, the CoopNavi case considers structured state information, trajectory data, and the initial spatial layout image of agents and landmarks. All methods operate on the same multimodal scenario input. NeSy-CSA constructs a structured attribution process via a validated predecessor-dependent subtask graph, where each intermediate result is explicitly grounded in either symbolic execution or dependency-constrained neural inference. The final attribution is therefore derived from an evidence chain rather than a direct end-to-end inference. The evidence graph reveals that most candidate factors, including trajectory smoothness, landmark approach rate, and collision-related checks, contribute marginally. The decisive factor is the final Agent 2-Landmark 3 distance (0.22), which exceeds the distance threshold of 0.2. Together with a high velocity synchronization coefficient, the result indicates that the critical outcome is primarily driven by insufficient progress of Agent 2 under coordinated multi-agent dynamics.

LLM-based baselines exhibit different failure modes. In LLM-only, attribution is generated in an end-to-end manner without intermediate grounding or structural constraints, which may produce hallucinated yet plausible explanations. The LLM-only model therefore incorrectly attributes the critical outcome to agent collisions, despite no supporting evidence in the trajectory. In LLM+CoT, the multi-step reasoning chain leads to error propagation across intermediate steps without validation. In this case, CoT incorrectly concludes that Agent 1 fails to reach its assigned landmark due to excessive initial distance. However, the trajectory evidence shows that Agent 1 successfully reaches its target, indicating that the final attribution is inconsistent with ground-truth behavior. This error arises because intermediate reasoning steps are not grounded in executable checks or dependency constraints, allowing early misinterpretations to persist and accumulate into an incorrect final conclusion. LLM+Tool enhances attribution by invoking external tools to improve local numerical computations. However, it does not perform factor-level restriction of the attribution space, and thus cannot constrain reasoning to the most discriminative causal factors. As a result, LLM+Tool incorporates weakly relevant interaction signals while missing key factors that determine the outcome, leading to an unbalanced attribution space where causal indicators are unevenly represented. Since it operates over a predefined and incomplete tool set, it cannot fully support the computations required for complete causal reasoning. Consequently, attribution is performed in a misaligned reasoning space without a traceable reasoning structure, leading to incomplete causal pathways and ultimately yielding a vague and unreliable attribution conclusion.


\section{Conclusion}

This paper addressed the problem of open-ended critical scenario attribution for decision-making agents. Different from existing methods that mainly focus on discovering critical scenarios or generating unconstrained textual explanations, we proposed NeSy-CSA to transform critical scenario attribution into a structured and traceable reasoning process. The proposed framework constrains attribution from three complementary perspectives: refining the attribution factor space, organizing the reasoning process through predecessor-dependent subtasks, and combining symbolic execution with neural inference. In addition, we introduced a two-level evaluation protocol to assess both the consistency of the reasoning workflow and the practical impact of the identified factors on task outcomes. Experiments across different decision-making environments demonstrate that NeSy-CSA can produce more traceable and intervention-effective attribution results, thereby turning discovered critical scenarios into reusable knowledge for subsequent testing and safety analysis.

Future work will integrate attribution results into scenario generation to form a closed-loop testing process. Attribution conclusions can provide directions for scenario mutation or generation by indicating critical elements, perturbation trends, and interaction patterns that deserve further exploration. The generated scenarios can then be tested and re-attributed, enabling iterative discovery, explanation, and reuse of critical scenarios for targeted safety evaluation.



\bibliographystyle{IEEEtran}
\small\bibliography{ref}
\end{document}